%% file: root.tex
\documentclass[conference]{IEEE_ITSC}  % Comment this line out if you need a4paper

\IEEEoverridecommandlockouts                              % This command is only needed if 
\input{packages}
\def\compileforpublish{1}
\def\isaccepted{1}
\input{macros}

\definecolor{mn}{RGB}{255,127,0}
\definecolor{ts}{RGB}{0,0,255}
\definecolor{tosch}{RGB}{190,0,80}

\renewcommand{\IEEEtitletopspaceextra}{20pt}

\title{Assessment of Deep Convolutional Neural Networks for Road Surface Classification}
\author{Marcus Nolte, Nikita Kister and Markus Maurer\\%
	\IEEEauthorblockA{Institute of Control Engineering\\
	Technische Universit\"at Braunschweig\\
	Braunschweig, Germany\\
	Email: \{nolte, maurer\}@ifr.ing.tu-bs.de}%
}

\begin{document}
\maketitle%
\thispagestyle{empty}%
\pagestyle{empty}%
\copyrightnotice%
%
%%%%%%%%%%%%%%%%%%%%%%%%%%%%%%%%%%%%%%%%%%%%%%%%%%%%%%%%%%%%%%%%%%%%%%%%%%%%%%%%
\begin{abstract}%
\input{00_abstract}%
\end{abstract}%
%
%%%%%%%%%%%%%%%%%%%%%%%%%%%%%%%%%%%%%%%%%%%%%%%%%%%%%%%%%%%%%%%%%%%%%%%%%%%%%%%%
%
\section{Introduction}
\label{sec:intro}
\input{01_introduction}

\section{Related Work}
\label{sec:related}
\input{02_related_work}

\section{Challenges Regarding Available Datasets}
\label{sec:challenges}
\input{03_challenges}
\section{Approach for Surface Classification}
\label{sec:approach}
\input{04_approach}
\section{Training Parameters}
\label{sec:architectures}
\input{05_architecture}

\section{Results}
\label{sec:results}
\input{06_results}
\section{Conclusion \& Future Work}
\label{sec:conclusion}
\input{07_conclusion}
\section*{Acknowledgement}
\label{sec:ack}
\input{08_acknowledgement}

\IEEEtriggeratref{1}
\renewcommand*{\bibfont}{\footnotesize} 
\printbibliography 
\end{document}

%% file: packages.tex
\usepackage{tabularx,calc}
\usepackage{color}
\usepackage{tubscolors}

\usepackage{amsmath}
\usepackage{tikz}
\usetikzlibrary{shapes,automata,arrows,matrix,backgrounds,fit,patterns,decorations.markings, svg.path, shapes.multipart, external, shadows, positioning, calc, decorations}
\usepackage{pgfplots}
\usepackage{pgfplotstable}
\usepgfplotslibrary{groupplots}

\usepackage{siunitx}
\usepackage[style=ieee,backend=bibtex,minnames=1,maxcitenames=2,maxbibnames=99,doi=false,isbn=false,url=false,natbib=true]{biblatex} 
\usepackage[font=footnotesize]{subcaption}
\usepackage{hyperref}
\usepackage{ifthen}
\usepackage{lipsum}

\usepackage{amssymb} 	% extended math symbols
\usepackage{bm} 		% bold math 
\usepackage{booktabs} 	% booktabs for nicer tables
\usepackage{multirow}   % columns spanning multiple rows
\usepackage{colortbl}
\usepackage{array}
\usepackage{environ}

\AtBeginDocument{} 
\AtBeginDocument{} 

\newcolumntype{P}[1]{>{\centering\arraybackslash}p{#1}}

%% file: macros.tex
\makeatletter
\newcounter{IEEE@bibentries}
\renewcommand\IEEEtriggeratref[1]{%
	\renewbibmacro{finentry}{%
		\stepcounter{IEEE@bibentries}%
		\ifthenelse{\equal{\value{IEEE@bibentries}}{#1}}
		{\finentry\@IEEEtriggercmd}
		{\finentry}%
	}%
}
\makeatother

\ifx \isaccepted \undefined
\newcommand\copyrighttext{%
	\footnotesize \centering This work has been submitted to the IEEE for possible publication.\\ Copyright may be transferred without notice, after which this version may no longer be accessible.}
\else
\newcommand\copyrighttext{%
	\footnotesize \parbox[t]{.11\textwidth}{\copyright{} \the\year~IEEE.} \parbox[t]{.89\textwidth}{Personal use of this material is permitted. Permission from IEEE must be obtained for all other uses, in any current or future media, including reprinting/republishing this material for advertising or promotional purposes, creating new collective works, for resale or redistribution to servers or lists, or reuse of any copyrighted component of this work in other works.}}
\fi
\newcommand\copyrightnotice{%
	\ifx \compileforpublish \undefined
	\else
	\begin{tikzpicture}[remember picture,overlay]
	\node[anchor=south,yshift=10.5pt] at (current page.south) {\parbox{\dimexpr\textwidth-\fboxsep-\fboxrule\relax}{\copyrighttext}};
	\end{tikzpicture}%
	\fi
}

\newcommand{\drawconfusion}[1]{%
    \pgfplotsset{
	width=5cm,
	height=5cm,
	compat=1.3,
	colormap={whiteblue}{color=(white) color=(tuDarkBlue)},
	xticklabels={concrete, dirt, grass, wet, cobble, snow},
	yticklabels={concrete, dirt, grass, wet, cobble, snow},
	xtick={0,...,6},
	ytick={0,...,6},
}
\begin{tikzpicture}[every node/.style={font=\footnotesize}]
\begin{axis}[
enlargelimits=false,
xlabel style={yshift=7},
xtick align=outside,
tick pos=left,
ytick align = outside,
xlabel={predicted labels},
ylabel style={yshift=-4},
ylabel={ground truth labels},
legend style={},
xticklabel style={ rotate=45, anchor=east, yshift=-5},
yticklabel style={},
colorbar,
colorbar style={
	width=0.3cm,
	xshift=-5,
	ytick style={draw=none},
	ytick={0.03,0.22,0.41,0.59,0.78,0.97},
	yticklabels={0.0,0.2,0.4,0.6,0.8,1.0},
	yticklabel={\pgfmathprintnumber\tick},
	yticklabel style={font=\sf\scriptsize, /pgf/number format/assume math mode, 	        
		/pgf/number format/.cd,
		fixed,
		fixed zerofill,
		precision=1},
},
point meta min=0,
point meta max=1,
nodes near coords={\pgfmathprintnumber\pgfplotspointmeta},
nodes near coords black white/.style={
	small value/.style={
		font={\sffamily\scriptsize},
		/pgf/number format/assume math mode,
		yshift=-7pt,
		text=black,
		/pgf/number format/.cd,
		fixed,
		fixed zerofill,
		precision=2,
		/tikz/.cd                
	},
	large value/.style={
		font={\sffamily\scriptsize},
		/pgf/number format/assume math mode,
		yshift=-7pt,
		text=white,
		/pgf/number format/.cd,
		fixed,
		fixed zerofill,
		precision=2,
	},
	every node near coord/.style={
		check for zero/.code={
			\begingroup
			\pgfkeys{/pgf/fpu}
			\pgfmathparse{\pgfplotspointmeta<#1}
			\global\let\result=\pgfmathresult
			\endgroup
			\pgfmathfloatcreate{1}{1.0}{0}
			\let\ONE=\pgfmathresult
			\ifx\result\ONE
			\pgfkeysalso{/pgfplots/small value}
			\else
			\pgfkeysalso{/pgfplots/large value}
			\fi
		},
		check for zero,
	},
},
nodes near coords black white=0.5,
]
\addplot[
matrix plot,
mesh/cols=6,
point meta=explicit,
point meta min = -0.2,
point meta max = 1.2
] table [meta=meta,col sep=semicolon] {figures/data/#1};
\end{axis}
\end{tikzpicture}
}

\newcolumntype{M}[1]{>{\centering\arraybackslash}m{#1}}

%% file: 00_abstract.tex
When parameterizing vehicle control algorithms for stability or trajectory control, the road-tire friction coefficient is an essential model parameter when it comes to control performance.
One major impact on the friction coefficient is the condition of the road surface.
A camera-based, forward-looking classification of the road-surface helps enabling an early parametrization of vehicle control algorithms.
In this paper, we train and compare two different Deep Convolutional Neural Network models, regarding their application for road friction estimation and describe the challenges for training the classifier in terms of available training data and the construction of suitable datasets.

%\begin{itemize}
%	\item knowledge of road conditions is important for proper parameterization of vehicle control algorithms
%	\item CNN great for classification
%	\item main challenge: available datasets not suitable for broad range of surfaces
%	\item idea: augment dataset with texture patches
%	\item paper analyses impact of augmentation \& mis-classifications
%\end{itemize}

%% file: 01_introduction.tex
%\begin{itemize}
%	\item fully automated vehicles require wide range of operational scenarios
%	\item road surface important factor for control algorithms
%	\item estimation of friction coefficient is holy grail of vehicle dynamics
%	\item many approaches measure road surface "under the vehicle" (slip comparision, radar, ...)
%	\item for early adaption of control strategy lookahead is important
%	\item camera for looking in front of vehicle
%	\item present CNN-based approach
%	\item compare InceptionV3 and ResNet50
%	\item perform data augmentation from web search
%	\item evaluate effects \& mis-classifications
%\end{itemize}
Systems for vehicle dynamics control have been implemented in series vehicles for several decades.
A central challenge for the implementation of well-performing control algorithms is the estimation of the road-tire friction coefficient $\mu$ which models the maximal adhesive force between the vehicle's tires and the road surface.
While its exact value depends on a variety of factors, such as tire and road temperatures and the composition of the tire, the road surface condition has major impact on the maximal transmittable drive or brake force.
Thus proper estimation of the friction coefficient is a widely discussed topic in the field of vehicle dynamics.

Many presented approaches (such as \cite{valada2017,han2017}) are of reactive nature, which means that e.g. the current measured vehicle dynamics are utilized in observer-based systems to estimate the friction coefficient.
An alternative for reactive estimation is the utilization of sensors under the vehicle (microphones, radar, optical sensors) which are used for recording the road surface under the vehicle.
While such reactive approaches have been shown to increase control performance \cite{han2017}, predictive approaches promise additional benefits for control performance, as a look-ahead estimation allows an early adoption of control algorithms to upcoming road conditions.
Also when going beyond vehicle control and regarding trajectory planning for automated vehicles, knowledge about the road conditions in front of the vehicle is beneficial, because the gained knowledge allows an adoption of planning strategies e.g. when approaching wet or snowy road patches.

While lidar and radar sensors allow detection of wet surfaces, due to different reflectivity, camera images provide high-resolution texture information.
Texture information does not only allow the detection of wet surfaces, but also allows to differentiate e.g. between concrete and cobblestone roads.
This additional information has already been exploited for predictive estimation of the road friction coefficient \cite{holzmann2006, omer2010, qian2016}.
As Deep Convolutional Neural Networks (CNNs) have been successfully applied to different classification tasks, also with applications in the field of automated driving, it seems promising to use a CNN-based approach for surface classification.

However, the performance of learned classifiers heavily relies on the design of training data.
Most available datasets are recorded in dry conditions in inner cities.
Hence, the datasets provide unbalanced class information for training a CNN for the given task.
We create a mixed dataset from publicly available datasets for automated driving (KITTI \cite{geiger2013}, Oxford Robocar Dataset \cite{maddern2017}), own recorded data from the project \emph{Stadtpilot} \cite{nothdurft2011a} as well as images from datasets not particularly designed for automated driving (\cite{maddern2017, giusti2016, smith2009}) as well as images from web search.
Based on this mixed dataset, this paper presents two different convolutional network architectures based on ResNet50 \cite{he2016} and InceptionNetV3 \cite{szegedy2016} to differentiate six classes of road surface conditions.
The results acquired from both networks are discussed with respect to the design of the training dataset.

The remainder of the paper is organized as follows:
\autoref{sec:related} summarizes previous work on predictive and reactive $\mu$-estimation. 
\autoref{sec:challenges} describes the challenge of creating balanced datasets for training a CNN for the given task.
After describing the general approach for classification in \autoref{sec:approach}, \autoref{sec:architectures} describes the training parameters for the evaluated network architectures.
The obtained results are discussed in \autoref{sec:results} before concluding with an outlook on future applications for the presented method (\autoref{sec:conclusion}).

%% file: 02_related_work.tex
%\begin{itemize}
%	\item \todo{some research, papers from BA}
%	\item Uwe Franke's stuff with semantic segmentation
%	\item ResNet
%	\item InceptionNet
%	\item maybe texture classification stuff?
%\end{itemize}
%\todo{0.5 - 0.75 S.}
%Approaches for estimating the road-tire friction coefficient can be divided into model-based and sensor-based approaches \cite{khaleghian2017}. 
%While the former can only be of reactive nature, the latter approaches can also be predictive, when utilizing sensors which provide information about the road surface in front of the vehicle.

Regarding related work, \citeauthor{khaleghian2017} present a literature review about recent publications for road friction estimation \cite{khaleghian2017}.
For this paper, we will concentrate on publications discussing camera-based surface classification and related learning-based applications.

A combined predictive approach for road surface classification from audio and video data is presented in \cite{holzmann2006}. 
They use luminance-based co-occurrence matrices to differentiate texture properties in parts of the image.
They differentiate six different classes (dry, humid, and wet asphalt, and cobblestone, respectively) of road surfaces in daylight and nighttime conditions.
The image-based classification is fused with audio information, which is processed to extract characteristic frequencies of the different road surfaces.
Unfortunately, no information about the performance of both methods is provided.

\citeauthor{omer2010} present a pure image-based approach for detecting snow-covered roads \cite{omer2010}.
A region of interest in front of the vehicle is used to train a support vector machine with partial image histograms as features.
Using geolocation information as additional features, an average classification accuracy of \SI{85}{\percent} could be achieved.

\citeauthor{qian2016} evaluate different learning-based approaches for surface classification in a dynamic region of interest \cite{qian2016}.
Evaluated features were MR8 \cite{varma2009} as well as SURF features \cite{bay2006}.
For classification, Fisher Vectors were compared to Bag of Visual Words as well as a Texton dictionary.
The classifiers were trained to perform binary classification (asphalt vs. snow- or ice-covered road) over three-class (dry vs. wet asphalt vs. snow- or ice-covered road) to five-class classification (dry vs. wet asphalt vs. snow covered vs. snow packed).
MR8 features with a bag of words classifier provided the best results with \SI{98}{\percent} average classification accuracy for the binary classification problem with a manually defined region of interest.
For the five-class problem, classification accuracy dropped to \SI{62}{\percent}.

\citeauthor{valada2017} present an audio-based approach and train a convolutional neural network with recurrent long short-term memory (LSTM) units to differentiate nine classes (asphalt, mowed grass, high grass, paving, cobblestone, dirt, wood, linoleum, and carpet) \cite{valada2017}.
Input data for the convolutional layers are spectrograms extracted via Short Term Fourier Transform.
The approach reaches an average classification accuracy of \SI{97.52}{\percent} (CNN only) and \SI{98.67}{\percent} (CNN+LSTM), respectively

%% file: 03_challenges.tex
A challenge for training deep neural networks is the availability of suitable, annotated training data.
One challenge for training neural networks for classification tasks is the class imbalance problem caused by over-represented classes (majority classes) and under-represented classes (minority classes) in a dataset:
If single classes dominate a training set or single classes are only represented by a small number of samples, classification performance can degrade significantly \cite{buda2017}.

For the application of deep convolutional networks to road surface classification this has the following consequences: 
While there are many datasets available for general image classification (ImageNet \cite{deng2009}) or autonomous driving in general, such as KITTI \cite{geiger2013}, a specific dataset for road surface classification is not available.
Resorting to those general datasets for automated driving results in heavily imbalanced datasets, as the majority of images was recorded in sunny or overcast weather, for reasons of better illumination and less optical obstructions due to rain on the windscreen.
Furthermore, the majority of recorded road surfaces is flat asphalt, while surfaces such as dirt road or sand are not represented, as they do not appear on city roads. 

\subsection{Composition of Dataset}
For the selection of suitable training data we thus looked at a variety of available datasets which should provide a more balanced set of images for surface types as a whole.
In addition, composing training data from multiple datasets has the advantage of covering several different cameras which helps avoiding learning features specific to the camera-setup of a single research vehicle. 
One restriction we applied was that the perspective towards the surface should be vaguely similar to the perspective of a windscreen mounted camera, to avoid applying artificial distortion of the chosen images. 
For the composition of the dataset we used images from the following datasets:

\begin{itemize}
	\item NREC Human Detection \& Tracking in Agriculture \cite{pezzementi2017}
	\item KITTI Vision Benchmark Suite \cite{geiger2013}
	\item Oxford Robocar Dataset \cite{maddern2017}
	\item New College Vision and Laser Data Set \cite{smith2009}
	\item Imageset published by \cite{giusti2016}
	\item Image sequences captured in the Stadtpilot project by our research vehicle \emph{Leonie} \cite{nothdurft2011a}
\end{itemize}

The obtained classes available in each individual dataset are presented in Table \ref{tab:datasets}.
\bgroup
\def\arraystretch{1.24}
\begin{table}
	\caption{Available classes in the individual datasets. Numbers in parentheses denote total number of selected samples.}
	\begin{tabularx}{\columnwidth}[c]{m{1.2cm}M{0.8cm}M{0.5cm}M{0.8cm}M{0.8cm}M{1cm}M{0.6cm}}\toprule
					   & asphalt\newline (10273) & dirt\newline (8547) & grass\newline (2887) & wet\newline asphalt \newline (3668) & cobble-\newline stone\newline(1082) & snow\newline(3075) \\ \midrule
		Robocar 	   & X 		 & 		& X 	& X 		  &  			& X \\
		\rowcolor[gray]{.9} Stadtpilot	   & X		 &	    & 	    & X	          & X			& \\
		NREC		   &  		 & X 	& X		& 			  & 			& \\
		\rowcolor[gray]{.9} New\newline College	   & X		 & X	& X 	& 			  & 			&  \\
		Giusti\newline  et al.  & X 		 & X 	& X 	& 			  & X			& X \\
		\rowcolor[gray]{.9} KIITI 		   & X		 & 		& X 	&			  & X 			& \\	\bottomrule	
	\end{tabularx}
	\label{tab:datasets}
\end{table}
\egroup

Analyzing these datasets and comparing the majority class asphalt with the minority class cobblestone yields an imbalance ratio of 10:1:
The class \emph{asphalt} consists of over 10000 images, while the class \emph{cobblestone} is represented in just over 1300 images.

To counteract the imbalance, instead of applying over or under sampling, we added further images from Google image search, following the example of \cite{krause2016} for fine grained image classification.

%An additional challenge for the task of road surface classification is that parts of the available datasets suffer from . 
%Some examples are shown in Figure \todo{bilder einfügen}.

\subsection{Selection of Test and Training Data}

All used datasets provide frame sequences rather than a random collection of independently recorded frames.
Thus the road conditions vary only slightly between frames from a single sequence.
When dividing the selected images into test and training sets, we did not only split single sequences, but also selected images from different sequences for test where possible.

The finally used test set consisted of 300 images per class.
The remaining images were used for training, building three different training sets.

A first set only consisted of the images from the datasets mentioned above.
To create a balanced set, 700 images per class were chosen randomly.
300 images were used for validation. 

For the second dataset, the classes \emph{cobblestone} and \emph{wet asphalt} were extended with images from Google image search as mentioned earlier. 
The class \emph{wet asphalt} was available in the fewest sequences, while the class \emph{cobblestone} had the lowest number of samples.
As the image search resulted in too few usable samples (\emph{grass}) or a sufficient number of images was available (\emph{asphalt}), only the first two classes were extended.
Using the Google image search, the class \emph{cobblestone} was extended such that each class consisted of 2500 images.
The class \emph{wet asphalt} was extended to increase variation of images within the class. 
The training set was thus more than doubled. 
500 images were selected for validation.

For a third dataset all classes from the basic dataset were extended with 300 images from Google image search, which corresponds to the number of returned usable images for the class \emph{grass}.
 
In order to overcome the issue of lacking variation between consecutive frames in the sequences, the used sequences were subsampled, using only every $n^{th}$ frame, with $n$ depending on the length of the sequence.

%% file: 04_approach.tex
\begin{figure}[!h]%CLASS WINDOW
\centering
  \begin{subfigure}{0.29\columnwidth}
    \centering
    \includegraphics[width=\textwidth, height=\textwidth]{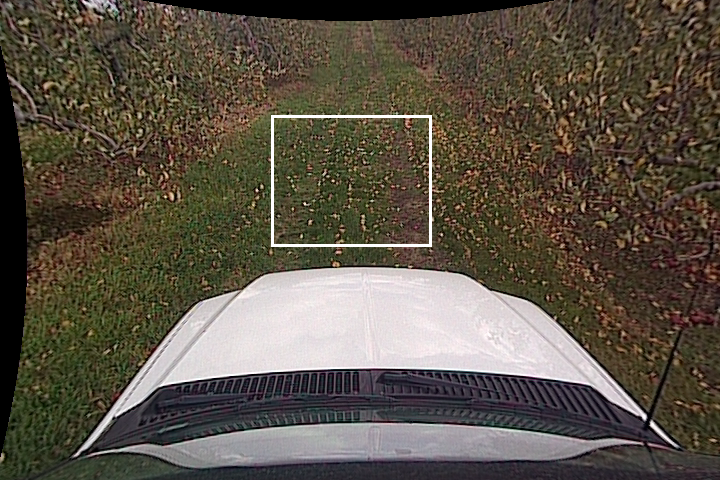}
  \end{subfigure}%
  \begin{subfigure}{0.29\columnwidth}
    \centering
    \includegraphics[width=\textwidth, height=\textwidth]{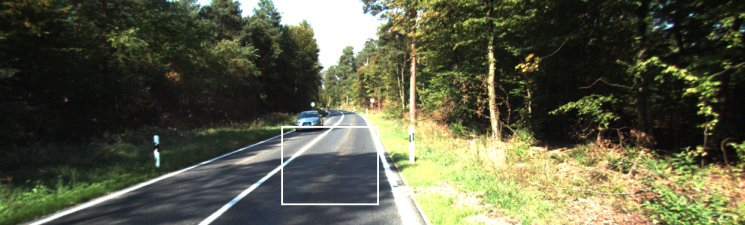}
  \end{subfigure}%
  \begin{subfigure}{0.29\columnwidth}
    \centering
    \includegraphics[width=\textwidth, height=\textwidth]{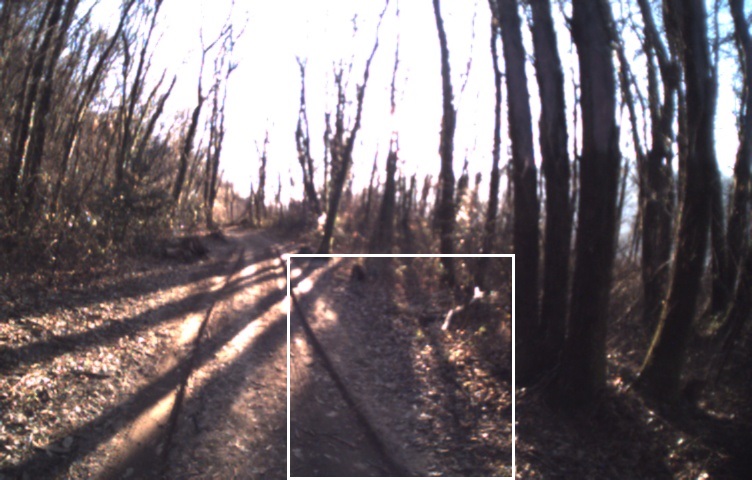}
  \end{subfigure}%
\\
  \begin{subfigure}{0.29\columnwidth}
    \centering
    \includegraphics[width=\textwidth, height=\textwidth]{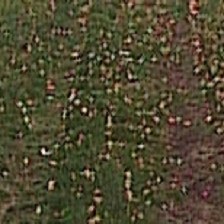}
  \end{subfigure}%
  \begin{subfigure}{0.29\columnwidth}
    \centering
    \includegraphics[width=\textwidth, height=\textwidth]{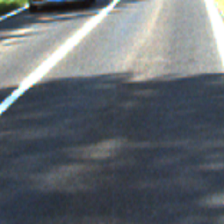}
  \end{subfigure}%
  \begin{subfigure}{0.29\columnwidth}
    \centering
    \includegraphics[width=\textwidth, height=\textwidth]{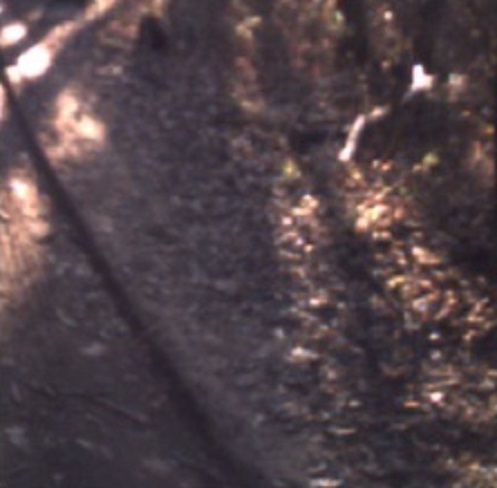}
  \end{subfigure}%
\caption{Some example images for the choice of the region of interest. The cropped images were resized to 224 by 224 px for the classification.}
\label{fig:window}
\end{figure}

%\begin{itemize}
%	\item goal: extract road surface conditions in front of vehicle
%	\item possible: whole image, image patch
%	\item define patch
%	\item use patches for training $\Rightarrow$ faster 
%	\item classify road surface in patch
%\end{itemize}
%\todo{0.5 - 0.75 S.}
% image which part?
% whole image leads to overfitting -> evidence
% use only the part which captures the road
% how to choose? -> road is most of the time in the same place 
% static window is sufficient
% BUT depends on the dataset 
%  different different perspective cause by different vehicle and different moutning
In order to classify the road conditions in front of the vehicle, several strategies were evaluated, including running the classification on the whole image and selecting regions of interest.
Performing the classification task on the whole image provides additional information about the environment, such as light conditions and the sky.
However, evaluations showed that this approach resulted in severe overfitting, providing a validation accuracy of ~\SI{80}{\percent}. 
Therefore, classification was performed on a region of interest (cf.~Fig.~\ref{fig:window}) which increased validation accuracy to over \SI{90}{\percent} as will be presented in the results section.

As the position of the road differs in each of the chosen data sets due to different cameras with different fields of view, the position of the region of interest was defined individually for each dataset.
The extracted texture patches were resized to 224 x 224 px for training and classification.

%% file: 05_architecture.tex
%\begin{itemize}
%	\item summarize network architecture
%	\item different approaches
%	\item pro's and con's
%\end{itemize}
%\todo{1.5 S.}
As mentioned above, we evaluated ResNet50 and InceptionV3 for the classification task.
Both architectures were initialized with pre-trained weights from the ImageNet dataset and trained using cross-entropy as a cost function minimized by stochastic gradient descent.
Batch normalization was applied.
The initial learning rates for both architectures were set to $3\cdot 10^{-5}$ in order to protect the pre-trained weights.
Training was performed with a batch size of 48.
Analysis of the validation accuracy showed no significant gain after five epochs, thus early stopping was applied in order to avoid overfitting

In order to account for the vehicle's motion and the resulting changes of perspective towards the road surface, we also applied data augmentation for each batch.
For this purpose the texture patches were mirrored horizontally, randomly rotated in an interval of $\pm$\SI{40}{\degree} and scaled by a random factor of 0.9 to 1.1.

As the regions of interest can contain ambiguous texture information, e.g. if multiple surface types are visible in the patch, the selected labels can be incorrect to a certain degree. 
For this reason we applied label smoothing with a factor of 0.1.

%% file: 06_results.tex
%\begin{itemize}
%	\item training performance
%	\item loss curves etc.
%	\item confusions matrices
%	\item exemplary sequences
%	\item analyze mis-classifications
%	\item try to explain why what has happened
%\end{itemize}
%\todo{1.5 S}
\subsection{Training \& Classification Performance}
This section presents an overview about the achieved training and classification results with both implemented architectures on the three datasets (basic, image search augmentation for two classes, augmentation for all classes) described above.
%
%\begin{figure}%INCEPTIONV3 TRAINING
%\centering
%  \begin{subfigure}{0.239\textwidth}
%    \resizebox{\textwidth}{!}{%
%      \drawaccuracy{figures/data/D1Inception.csv}
%    }
%    \caption{Basic dataset without augmentation}
%  \end{subfigure}
%%  
%  \begin{subfigure}{0.239\textwidth}
%    \resizebox{\textwidth}{!}{%
%      \drawaccuracy{figures/data/D2Inception.csv}    }
%    \caption{Dataset augmented with image search for class cobblestone}
%  \end{subfigure} \\
%\vspace{0.5em}
%  \begin{subfigure}{0.239\textwidth}
%    \resizebox{\textwidth}{!}{%
%      \drawaccuracy{figures/data/D3Inception.csv}    }
%    \caption{Dataset augmented with image search for all classes.}
%  \end{subfigure}
%%
%  \caption{InceptionV3 architecture trained on the three datasets.}
%  \label{val:acc:iv3}
%\end{figure}

\begin{figure}[bhtp]%RESNET50 TRAINING
	\centering
	\includegraphics[width=\columnwidth]{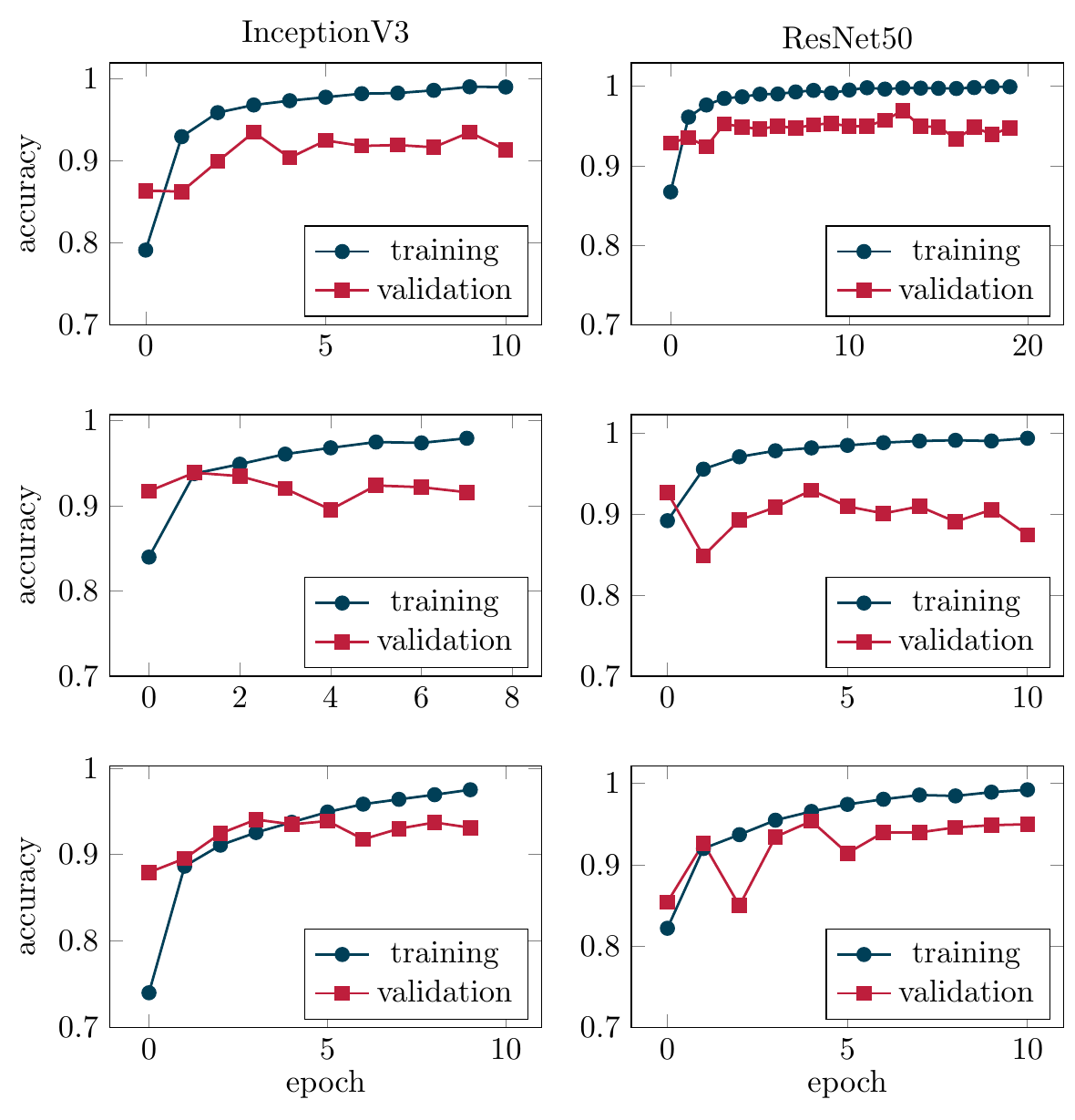}
	\caption{Training and validation accuracy of InceptionV3 (left) and ResNet50 (right) architectures trained on the three datasets. Top to bottom: basic dataset, dataset with class \emph{cobblestone} and \emph{wet asphalt} extended from image search, dataset with all classes augmented from image search. All data is plotted until training terminated due to early stopping.}
	\label{val:acc}
\end{figure}

Considering training performance, the InceptionV3 model terminates after seven (second training dataset) to ten (first training dataset) epochs.
The maximum validation accuracy is reached after the third epoch as shown in Figure \ref{val:acc} (left hand side). 
The average validation accuracies of the models trained on the basic dataset and the second dataset are comparable.

Extending the basic dataset with images from Google image search for all classes leads to a decrease of $\approx$\SI{1.5}{\percent} in validation accuracy on training data.
The extension of the training dataset does not impact the duration of the training.
The inference run on an NVIDIA Titan X GPU takes \SI{153}{\milli\second} in average. 

When evaluating the performance on the test dataset, the InceptionV3 architecture behaves differently:
Training the model on the first and second dataset resulted in a comparable test accuracy of \SI{90}{\percent}.
Extending all classes with images from image search, however, resulted in a an test accuracy of only \SI{84}{\percent}.
The behavior of the model provides a hint, that the additional variation of training data does not provide any benefit.
In contrary, the network starts to overfit

Training of the ResNet50 architecture takes longer than the training of the InceptionV3 model:
Training terminates after ten (basic dataset) to twenty epochs (second and third dataset).
In contrast to the InceptionV3 architecture, adding images from Google image search, speeds up the training process (cf.~Fig.\ref{val:acc}, right hand side).

The ResNet model trained on the first dataset achieved a lower test accuracy on the test dataset (\SI{80}{\percent}) than the  corresponding InceptionV3 model.
However, the partial addition of images from google image search increased the test accuracy by \SI{4}{\percent} to an overall \SI{92}{\percent}.

The ResNet50 architecture exposes the same over-fitting behavior as InceptionV3 with performance decreased to \SI{84}{\percent} when the basic dataset is extended with images for each class.
Inference for ResNet50 takes \SI{94}{\milli\second} on the NVIDIA Titan X.

\subsection{Analysis of Results}
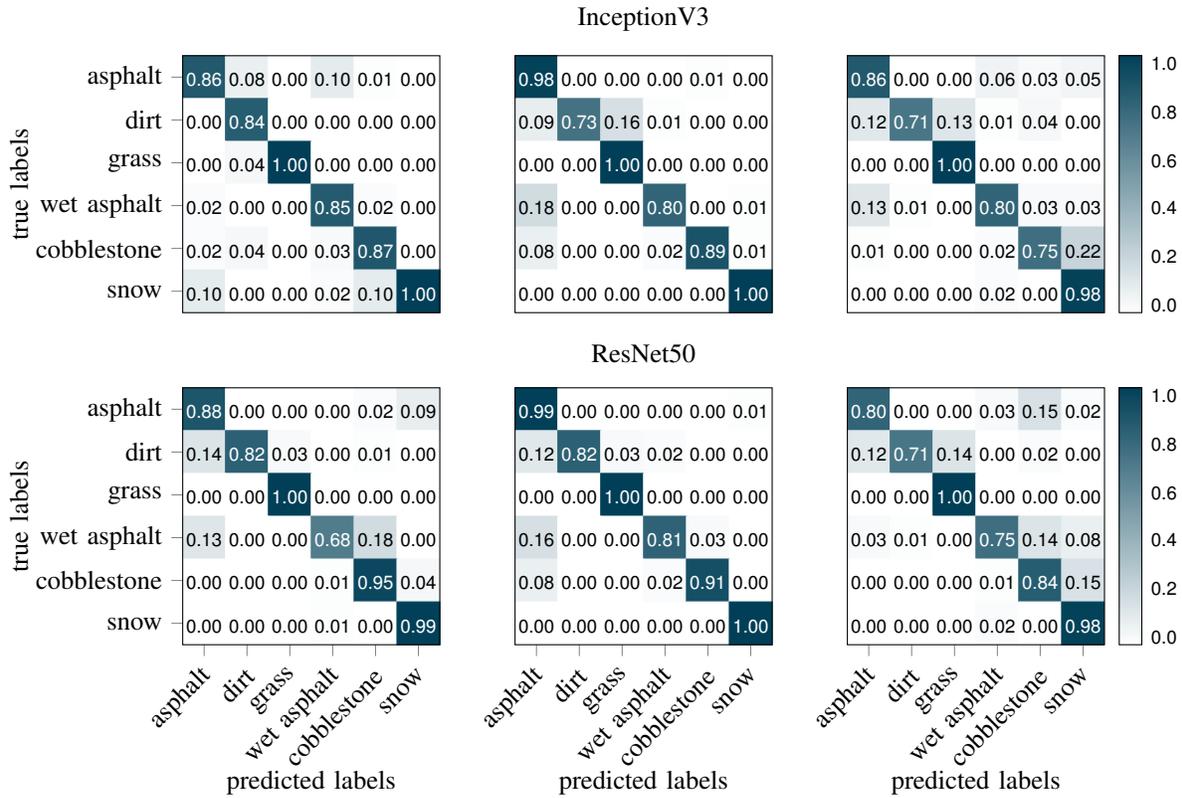
\begin{figure*}
	\centering
	\input{figures/tikz/plots_confusion.tikz}
	\caption{Evaluation on test data: Confusion matrices for trained InceptionV3 (top) and ResNet50 (bottom) architectures. Left to right: Basic dataset, dataset with classes \emph{cobblestone} and \emph{wet asphalt} extended from image search, dataset with all classes augmented from image search.}
	\label{fig:confusion}
	\vspace{-1em}
\end{figure*}

Looking at the confusion matrices in Figure \ref{fig:confusion},  misclassification occurs in a pattern. 
Images from the class \emph{wet asphalt} are often classified as \emph{asphalt}, but in no case with the class \emph{grass}, when augmenting the classes \emph{cobblestone} and \emph{wet asphalt}. 
In contrast classification accuracy for the class \emph{cobblestone} increases by \SI{2}{\percent} and for \emph{asphalt} by \SI{12}{\percent}.
Images from the classes \emph{snow} and \emph{grass} are classified with high recall. 

By examining the misclassified images, several possible causes for the misclassification could be identified.

\begin{figure}%color ambiguity
\centering
\begin{subfigure}{0.15\textwidth}
\includegraphics[width=\textwidth]{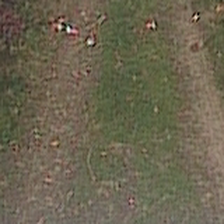}
\caption{Classified as "grass".}
\end{subfigure}
\begin{subfigure}{0.15\textwidth}
\includegraphics[width=\textwidth]{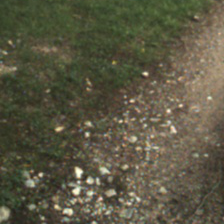}
\caption{Classified as "grass".}
\end{subfigure}
\begin{subfigure}{0.15\textwidth}
\includegraphics[width=\textwidth]{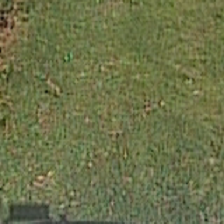}
\caption{Training image for the class "grass".}
\end{subfigure}
\begin{subfigure}{0.15\textwidth}
\includegraphics[width=\textwidth]{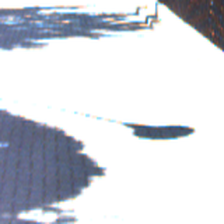}
\caption{Classified as "snow"}
\end{subfigure}
\begin{subfigure}{0.15\textwidth}
\includegraphics[width=\textwidth]{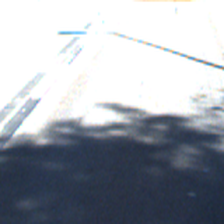}
\caption{Classified as "snow"}
\end{subfigure}
\begin{subfigure}{0.15\textwidth}
\includegraphics[width=\textwidth]{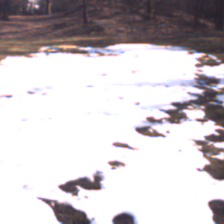}
\caption{Training image for the class \emph{snow}.}
\end{subfigure}
\caption{The first two images in each row were misclassified. The rightmost images were part of the training set.}
\label{miss:color}
\vspace{-1em}
\end{figure}

The first one is the dominating color in the images.  
Within the given classes \emph{snow} and \emph{grass}, the most distinctive feature is color, as grass is commonly green and a road covered with snow is commonly white. 
Color as a learned feature can thus result in a high recall score for both classes. 
Evaluating the false positive classifications, the resulting images consist of images which contain these colors.
Samples are shown in Figure \ref{miss:color}.
\begin{figure}[b]%color ambiguity
	\centering
	\begin{subfigure}{0.2\textwidth}
		\includegraphics[width=\textwidth]{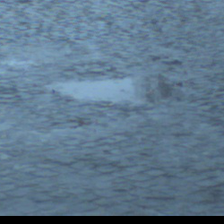}
	\end{subfigure}
	\begin{subfigure}{0.2\textwidth}
		\includegraphics[width=\textwidth]{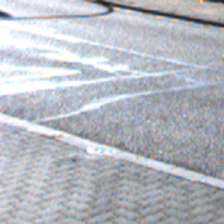}
	\end{subfigure}
	\caption{Examples for images with ambiguous classes: remaining puddles (left), cobblestone and asphalt in the same ROI (right)}
	\label{miss:ambig}
\end{figure}

When evaluating images from the class \emph{dirt}, the misclassified samples partially contain patches of grass, which makes the class prone to be misclassified as \emph{grass}. 
The same reason for misclassification applies to images containing overexposed regions which appear white and can thus be misclassified as \emph{snow}.
Not only the samples containing the class \emph{dirt} are prone to ambiguous texture information, as shown in Figure \ref{miss:ambig}. 

If the region of interest contains transitions between two road surfaces (e.g. \emph{asphalt} and \emph{cobblestone}, as shown on the right in Figure \ref{miss:ambig}), it contains features of two classes and are therefore also prone to misclassification. 
Another example for this are left-over puddles on an already dry cobblestone road (Fig. \ref{miss:ambig} on the left). 
Although part of the surface is wet, the surface should be considered as \emph{cobblestone}.

Although the classifier operates on single frames, the images are part of sequences.
In order to get an impression of the stability of the classification results when applied to those sequences, we evaluated the classification on sample sequences from the \emph{Stadtpilot} project, which were not part of the training dataset.
No tracking was performed between frames.
 
\begin{figure}%sequences
	\centering
	\includegraphics[width=.99\columnwidth]{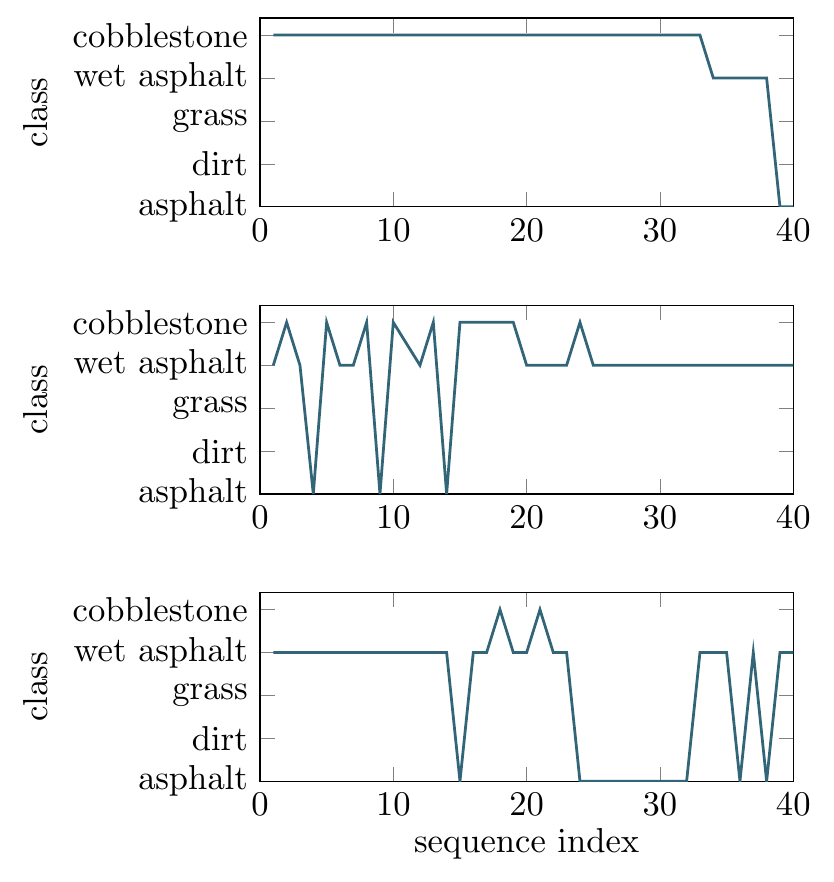}
	\caption{Classification results in sequences. Each image is classified separately, no tracking is performed. Ground truth from top to bottom: cobblestone, wet asphalt, wet asphalt.}
	\label{seq}
\end{figure}

For this classification, ResNet50 trained on the second dataset was used. 
In Figure \ref{seq} three of the worst classification results in sequences are shown.
Looking at these results, it is visible that misclassification tends to appear in groups of several frames. 
Fluctuations as shown in the center and bottom sequence can render the trained classifiers unsuitable for adapting control algorithms.

%% file: figures/tikz/plots_confusion.tikz
    \pgfplotsset{
        width=5cm,
        height=5cm,
        compat=1.3,
        colormap={whiteblue}{color=(white) color=(tuDarkBlue)},
        xticklabels={asphalt, dirt, grass, wet asphalt, cobblestone, snow},
        yticklabels={asphalt, dirt, grass, wet asphalt, cobblestone, snow},
        xtick={0,...,6},
        ytick={0,...,6},
    }
    
\begin{tikzpicture}
    \begin{groupplot}[group style={
                      group name=myplot,
                      group size= 3 by 2},height=5cm,width=5cm,
                      every plot/.style= {
                      ymin=0, 
		         x tick label style={/pgf/number format/1000 sep=},  
		         y tick label style={/pgf/number format/1000 sep=}, 
         }]
        \nextgroupplot[	%title=normalized confusion matrix,
        enlargelimits=false,
        xlabel style={yshift=7},
        xtick align=outside,
        tick pos=left,
        ytick align = outside,
        xlabel={},
        ylabel style={yshift=-4},
        ylabel={true labels},
        legend style={},
        xticklabel style={ rotate=45, anchor=east, yshift=-5},
        yticklabel style={},
        xticklabels={},
        xtick style={draw=none},
        colorbar style={
            width=0.3cm,
            xshift=-5,
            ytick style={draw=none},
            ytick={0.03,0.22,0.41,0.59,0.78,0.97},
            yticklabels={0.0,0.2,0.4,0.6,0.8,1.0},
            yticklabel={\pgfmathprintnumber\tick},
            yticklabel style={font=\sf\scriptsize, /pgf/number format/assume math mode, 	        
            	/pgf/number format/.cd,
	            fixed,
	            fixed zerofill,
	            precision=1},
        },
        point meta min=0,
        point meta max=1,
        nodes near coords={\pgfmathprintnumber\pgfplotspointmeta},
        % ---------------------------------------------------------------------
        % show `nodes near coords' but adapt the style so that values
        % above a threshold get another style
        % (adapted from <http://tex.stackexchange.com/a/141006/95441>)
        % #1: the THRESHOLD after which we switch to a special display.
        nodes near coords black white/.style={
            % define the style of the nodes with "small" values
            small value/.style={
                font={\sffamily\scriptsize},
                /pgf/number format/assume math mode,
                yshift=-7pt,
                text=black,
	        /pgf/number format/.cd,
	            fixed,
	            fixed zerofill,
	            precision=2,
	        /tikz/.cd                
            },
            % define the style of the nodes with "large" values
            large value/.style={
                font={\sffamily\scriptsize},
                /pgf/number format/assume math mode,
                yshift=-7pt,
                text=white,
                   /pgf/number format/.cd,
	            fixed,
	            fixed zerofill,
	            precision=2,
            },
            every node near coord/.style={
                check for zero/.code={
                        \begingroup
                        % this group is merely to switch to FPU locally.
                        % Might be unnecessary, but who knows.
                        \pgfkeys{/pgf/fpu}
                        \pgfmathparse{\pgfplotspointmeta<#1}
                        \global\let\result=\pgfmathresult
                        \endgroup
                        %
                        % simplifies debugging:
                        %\show\result
                        %
                        \pgfmathfloatcreate{1}{1.0}{0}
                        \let\ONE=\pgfmathresult
                        \ifx\result\ONE
                            % AH: our condition 'y < #1' is met.
                            \pgfkeysalso{/pgfplots/small value}
                        \else
                            % ok, proceed as usual.
                            \pgfkeysalso{/pgfplots/large value}
                        \fi
                },
                check for zero,
            },
        },
        % asign a value to the new style thich is the threshold at which
        % the two style `small value' or `large value' are used
        nodes near coords black white=0.5,
        % -----------------------------------------------------------------
    ]
        \addplot[
            matrix plot,
            mesh/cols=6,
            point meta=explicit,
            point meta min = -0.2,
            point meta max = 1.2
        ] table [meta=meta, col sep=semicolon] {figures/data/confusion.csv};
        
   \nextgroupplot[	%title=normalized confusion matrix,
        enlargelimits=false,
        xlabel style={yshift=7},
        xtick align=outside,
        tick pos=left,
        ytick align = outside,
        xlabel={},
        ylabel style={yshift=-4},
        ylabel={},
        legend style={},
        xticklabel style={ rotate=45, anchor=east, yshift=-5},
        yticklabel style={},
        yticklabels={},
        xticklabels={},
        ytick style={draw=none},
        xtick style={draw=none},
        title={InceptionV3},
        colorbar style={
            width=0.3cm,
            xshift=-5,
            ytick style={draw=none},
            ytick={0.03,0.22,0.41,0.59,0.78,0.97},
            yticklabels={0.0,0.2,0.4,0.6,0.8,1.0},
            yticklabel={\pgfmathprintnumber\tick},
            yticklabel style={font=\sf\scriptsize, /pgf/number format/assume math mode, 	        
            	/pgf/number format/.cd,
	            fixed,
	            fixed zerofill,
	            precision=1},
        },
        point meta min=0,
        point meta max=1,
        nodes near coords={\pgfmathprintnumber\pgfplotspointmeta},
        % ---------------------------------------------------------------------
        % show `nodes near coords' but adapt the style so that values
        % above a threshold get another style
        % (adapted from <http://tex.stackexchange.com/a/141006/95441>)
        % #1: the THRESHOLD after which we switch to a special display.
        nodes near coords black white/.style={
            % define the style of the nodes with "small" values
            small value/.style={
                font={\sffamily\scriptsize},
                /pgf/number format/assume math mode,
                yshift=-7pt,
                text=black,
	        /pgf/number format/.cd,
	            fixed,
	            fixed zerofill,
	            precision=2,
	        /tikz/.cd                
            },
            % define the style of the nodes with "large" values
            large value/.style={
                font={\sffamily\scriptsize},
                /pgf/number format/assume math mode,
                yshift=-7pt,
                text=white,
                   /pgf/number format/.cd,
	            fixed,
	            fixed zerofill,
	            precision=2,
            },
            every node near coord/.style={
                check for zero/.code={
                        \begingroup
                        % this group is merely to switch to FPU locally.
                        % Might be unnecessary, but who knows.
                        \pgfkeys{/pgf/fpu}
                        \pgfmathparse{\pgfplotspointmeta<#1}
                        \global\let\result=\pgfmathresult
                        \endgroup
                        %
                        % simplifies debugging:
                        %\show\result
                        %
                        \pgfmathfloatcreate{1}{1.0}{0}
                        \let\ONE=\pgfmathresult
                        \ifx\result\ONE
                            % AH: our condition 'y < #1' is met.
                            \pgfkeysalso{/pgfplots/small value}
                        \else
                            % ok, proceed as usual.
                            \pgfkeysalso{/pgfplots/large value}
                        \fi
                },
                check for zero,
            },
        },
        % asign a value to the new style thich is the threshold at which
        % the two style `small value' or `large value' are used
        nodes near coords black white=0.5,
        % -----------------------------------------------------------------
    ]
        \addplot[
            matrix plot,
            mesh/cols=6,
            point meta=explicit,
            point meta min = -0.2,
            point meta max = 1.2
        ] table [meta=meta, col sep=semicolon] {figures/data/inception_2.csv};        
        
   \nextgroupplot[	%title=normalized confusion matrix,
        enlargelimits=false,
        xlabel style={yshift=7},
        xtick align=outside,
        tick pos=left,
        ytick align = outside,
        ylabel style={yshift=-4},
        ylabel={},
        legend style={},
        xticklabel style={ rotate=45, anchor=east, yshift=-5},
        yticklabel style={},
        yticklabels={},
        xticklabels={},
        ytick style={draw=none},
        xtick style={draw=none},
        colorbar,
        colorbar style={
            width=0.3cm,
            xshift=-5,
            ytick style={draw=none},
            ytick={0.03,0.22,0.41,0.59,0.78,0.97},
            yticklabels={0.0,0.2,0.4,0.6,0.8,1.0},
            yticklabel={\pgfmathprintnumber\tick},
            yticklabel style={font=\sf\scriptsize, /pgf/number format/assume math mode, 	        
            	/pgf/number format/.cd,
	            fixed,
	            fixed zerofill,
	            precision=1},
        },
        point meta min=0,
        point meta max=1,
        nodes near coords={\pgfmathprintnumber\pgfplotspointmeta},
        % ---------------------------------------------------------------------
        % show `nodes near coords' but adapt the style so that values
        % above a threshold get another style
        % (adapted from <http://tex.stackexchange.com/a/141006/95441>)
        % #1: the THRESHOLD after which we switch to a special display.
        nodes near coords black white/.style={
            % define the style of the nodes with "small" values
            small value/.style={
                font={\sffamily\scriptsize},
                /pgf/number format/assume math mode,
                yshift=-7pt,
                text=black,
	        /pgf/number format/.cd,
	            fixed,
	            fixed zerofill,
	            precision=2,
	        /tikz/.cd                
            },
            % define the style of the nodes with "large" values
            large value/.style={
                font={\sffamily\scriptsize},
                /pgf/number format/assume math mode,
                yshift=-7pt,
                text=white,
                   /pgf/number format/.cd,
	            fixed,
	            fixed zerofill,
	            precision=2,
            },
            every node near coord/.style={
                check for zero/.code={
                        \begingroup
                        % this group is merely to switch to FPU locally.
                        % Might be unnecessary, but who knows.
                        \pgfkeys{/pgf/fpu}
                        \pgfmathparse{\pgfplotspointmeta<#1}
                        \global\let\result=\pgfmathresult
                        \endgroup
                        %
                        % simplifies debugging:
                        %\show\result
                        %
                        \pgfmathfloatcreate{1}{1.0}{0}
                        \let\ONE=\pgfmathresult
                        \ifx\result\ONE
                            % AH: our condition 'y < #1' is met.
                            \pgfkeysalso{/pgfplots/small value}
                        \else
                            % ok, proceed as usual.
                            \pgfkeysalso{/pgfplots/large value}
                        \fi
                },
                check for zero,
            },
        },
        % asign a value to the new style thich is the threshold at which
        % the two style `small value' or `large value' are used
        nodes near coords black white=0.5,
        % -----------------------------------------------------------------
    ]
        \addplot[
            matrix plot,
            mesh/cols=6,
            point meta=explicit,
            point meta min = -0.2,
            point meta max = 1.2
        ] table [meta=meta, col sep=semicolon] {figures/data/inception_3.csv};        

        \nextgroupplot[	%title=normalized confusion matrix,
        enlargelimits=false,
        xlabel style={yshift=7},
        xtick align=outside,
        tick pos=left,
        ytick align = outside,
        xlabel={predicted labels},
        ylabel style={yshift=-4},
        ylabel={true labels},
        legend style={},
        xticklabel style={ rotate=45, anchor=east, yshift=-5},
        yticklabel style={},
        colorbar style={
            width=0.3cm,
            xshift=-5,
            ytick style={draw=none},
            ytick={0.03,0.22,0.41,0.59,0.78,0.97},
            yticklabels={0.0,0.2,0.4,0.6,0.8,1.0},
            yticklabel={\pgfmathprintnumber\tick},
            yticklabel style={font=\sf\scriptsize, /pgf/number format/assume math mode, 	        
            	/pgf/number format/.cd,
	            fixed,
	            fixed zerofill,
	            precision=1},
        },
        point meta min=0,
        point meta max=1,
        nodes near coords={\pgfmathprintnumber\pgfplotspointmeta},
        % ---------------------------------------------------------------------
        % show `nodes near coords' but adapt the style so that values
        % above a threshold get another style
        % (adapted from <http://tex.stackexchange.com/a/141006/95441>)
        % #1: the THRESHOLD after which we switch to a special display.
        nodes near coords black white/.style={
            % define the style of the nodes with "small" values
            small value/.style={
                font={\sffamily\scriptsize},
                /pgf/number format/assume math mode,
                yshift=-7pt,
                text=black,
	        /pgf/number format/.cd,
	            fixed,
	            fixed zerofill,
	            precision=2,
	        /tikz/.cd                
            },
            % define the style of the nodes with "large" values
            large value/.style={
                font={\sffamily\scriptsize},
                /pgf/number format/assume math mode,
                yshift=-7pt,
                text=white,
                   /pgf/number format/.cd,
	            fixed,
	            fixed zerofill,
	            precision=2,
            },
            every node near coord/.style={
                check for zero/.code={
                        \begingroup
                        % this group is merely to switch to FPU locally.
                        % Might be unnecessary, but who knows.
                        \pgfkeys{/pgf/fpu}
                        \pgfmathparse{\pgfplotspointmeta<#1}
                        \global\let\result=\pgfmathresult
                        \endgroup
                        %
                        % simplifies debugging:
                        %\show\result
                        %
                        \pgfmathfloatcreate{1}{1.0}{0}
                        \let\ONE=\pgfmathresult
                        \ifx\result\ONE
                            % AH: our condition 'y < #1' is met.
                            \pgfkeysalso{/pgfplots/small value}
                        \else
                            % ok, proceed as usual.
                            \pgfkeysalso{/pgfplots/large value}
                        \fi
                },
                check for zero,
            },
        },
        % asign a value to the new style thich is the threshold at which
        % the two style `small value' or `large value' are used
        nodes near coords black white=0.5,
        % -----------------------------------------------------------------
    ]
        \addplot[
            matrix plot,
            mesh/cols=6,
            point meta=explicit,
            point meta min = -0.2,
            point meta max = 1.2
        ] table [meta=meta, col sep=semicolon] {figures/data/resnet_1.csv};
        
   \nextgroupplot[	%title=normalized confusion matrix,
        enlargelimits=false,
        xlabel style={yshift=7},
        xtick align=outside,
        tick pos=left,
        ytick align = outside,
        ylabel style={yshift=-4},
        ylabel={},
        legend style={},
        xticklabel style={ rotate=45, anchor=east, yshift=-5},
        yticklabel style={},
        yticklabels={},
        xlabel={predicted labels},
        ytick style={draw=none},
        title={ResNet50},
        colorbar style={
            width=0.3cm,
            xshift=-5,
            ytick style={draw=none},
            ytick={0.03,0.22,0.41,0.59,0.78,0.97},
            yticklabels={0.0,0.2,0.4,0.6,0.8,1.0},
            yticklabel={\pgfmathprintnumber\tick},
            yticklabel style={font=\sf\scriptsize, /pgf/number format/assume math mode, 	        
            	/pgf/number format/.cd,
	            fixed,
	            fixed zerofill,
	            precision=1},
        },
        point meta min=0,
        point meta max=1,
        nodes near coords={\pgfmathprintnumber\pgfplotspointmeta},
        % ---------------------------------------------------------------------
        % show `nodes near coords' but adapt the style so that values
        % above a threshold get another style
        % (adapted from <http://tex.stackexchange.com/a/141006/95441>)
        % #1: the THRESHOLD after which we switch to a special display.
        nodes near coords black white/.style={
            % define the style of the nodes with "small" values
            small value/.style={
                font={\sffamily\scriptsize},
                /pgf/number format/assume math mode,
                yshift=-7pt,
                text=black,
	        /pgf/number format/.cd,
	            fixed,
	            fixed zerofill,
	            precision=2,
	        /tikz/.cd                
            },
            % define the style of the nodes with "large" values
            large value/.style={
                font={\sffamily\scriptsize},
                /pgf/number format/assume math mode,
                yshift=-7pt,
                text=white,
                   /pgf/number format/.cd,
	            fixed,
	            fixed zerofill,
	            precision=2,
            },
            every node near coord/.style={
                check for zero/.code={
                        \begingroup
                        % this group is merely to switch to FPU locally.
                        % Might be unnecessary, but who knows.
                        \pgfkeys{/pgf/fpu}
                        \pgfmathparse{\pgfplotspointmeta<#1}
                        \global\let\result=\pgfmathresult
                        \endgroup
                        %
                        % simplifies debugging:
                        %\show\result
                        %
                        \pgfmathfloatcreate{1}{1.0}{0}
                        \let\ONE=\pgfmathresult
                        \ifx\result\ONE
                            % AH: our condition 'y < #1' is met.
                            \pgfkeysalso{/pgfplots/small value}
                        \else
                            % ok, proceed as usual.
                            \pgfkeysalso{/pgfplots/large value}
                        \fi
                },
                check for zero,
            },
        },
        % asign a value to the new style thich is the threshold at which
        % the two style `small value' or `large value' are used
        nodes near coords black white=0.5,
        % -----------------------------------------------------------------
    ]
        \addplot[
            matrix plot,
            mesh/cols=6,
            point meta=explicit,
            point meta min = -0.2,
            point meta max = 1.2
        ] table [meta=meta, col sep=semicolon] {figures/data/resnet_2.csv};        
        
   \nextgroupplot[	%title=normalized confusion matrix,
        enlargelimits=false,
        xlabel style={yshift=7},
        xtick align=outside,
        tick pos=left,
        ytick align = outside,
        ylabel style={yshift=-4},
        ylabel={},
        xlabel={predicted labels},
        legend style={},
        xticklabel style={ rotate=45, anchor=east, yshift=-5},
        yticklabel style={},
        yticklabels={},
        ytick style={draw=none},
        colorbar,
        colorbar style={
            width=0.3cm,
            xshift=-5,
            ytick style={draw=none},
            ytick={0.03,0.22,0.41,0.59,0.78,0.97},
            yticklabels={0.0,0.2,0.4,0.6,0.8,1.0},
            yticklabel={\pgfmathprintnumber\tick},
            yticklabel style={font=\sf\scriptsize, /pgf/number format/assume math mode, 	        
            	/pgf/number format/.cd,
	            fixed,
	            fixed zerofill,
	            precision=1},
        },
        point meta min=0,
        point meta max=1,
        nodes near coords={\pgfmathprintnumber\pgfplotspointmeta},
        % ---------------------------------------------------------------------
        % show `nodes near coords' but adapt the style so that values
        % above a threshold get another style
        % (adapted from <http://tex.stackexchange.com/a/141006/95441>)
        % #1: the THRESHOLD after which we switch to a special display.
        nodes near coords black white/.style={
            % define the style of the nodes with "small" values
            small value/.style={
                font={\sffamily\scriptsize},
                /pgf/number format/assume math mode,
                yshift=-7pt,
                text=black,
	        /pgf/number format/.cd,
	            fixed,
	            fixed zerofill,
	            precision=2,
	        /tikz/.cd                
            },
            % define the style of the nodes with "large" values
            large value/.style={
                font={\sffamily\scriptsize},
                /pgf/number format/assume math mode,
                yshift=-7pt,
                text=white,
                   /pgf/number format/.cd,
	            fixed,
	            fixed zerofill,
	            precision=2,
            },
            every node near coord/.style={
                check for zero/.code={
                        \begingroup
                        % this group is merely to switch to FPU locally.
                        % Might be unnecessary, but who knows.
                        \pgfkeys{/pgf/fpu}
                        \pgfmathparse{\pgfplotspointmeta<#1}
                        \global\let\result=\pgfmathresult
                        \endgroup
                        %
                        % simplifies debugging:
                        %\show\result
                        %
                        \pgfmathfloatcreate{1}{1.0}{0}
                        \let\ONE=\pgfmathresult
                        \ifx\result\ONE
                            % AH: our condition 'y < #1' is met.
                            \pgfkeysalso{/pgfplots/small value}
                        \else
                            % ok, proceed as usual.
                            \pgfkeysalso{/pgfplots/large value}
                        \fi
                },
                check for zero,
            },
        },
        % asign a value to the new style thich is the threshold at which
        % the two style `small value' or `large value' are used
        nodes near coords black white=0.5,
        % -----------------------------------------------------------------
    ]
        \addplot[
            matrix plot,
            mesh/cols=6,
            point meta=explicit,
            point meta min = -0.2,
            point meta max = 1.2
        ] table [meta=meta, col sep=semicolon] {figures/data/resnet_3.csv};  
    \end{groupplot}

\end{tikzpicture}

%% file: 07_conclusion.tex
%\begin{itemize}
%	\item image based classification can work
%	\item augmentation works, best with ResNet50, increases accuracy about \SI{4}{\percent}
%	\item patch based outperforms image based
%	\item consequence: use presegmented data
%	\item local surfaces could be used
%	\item difficult annotation
%\end{itemize}
%\todo{0.25 - 0.5 S.}
In this paper we presented an approach for CNN-based road surface classification, which can be used as a basis to predict the road friction coefficient.
The trained network models are able to differentiate between six types of surface labels.
Augmenting data from publicly available datasets for automated driving with images from Google search for minority classes has helped to increase the overall classification accuracy of ResNet50 by \SI{4}{\percent} and reduced confusion when differentiating between wet asphalt and cobblestone.

For the presented task, the trained ResNet50 model outperforms InceptionV3 by \SI{2}{\percent} with respect to average classification accuracy on test data, if the basic dataset is extended with additional images for minority classes.
With an average accuracy of \SI{92}{\percent} on the test dataset, the approach performs better than the image-based approaches using classic features and classifiers \cite{qian2016, omer2010}.
Unfortunately neither \cite{omer2010} nor \cite{qian2016} present more detailed results about precision and recall, such that a comparison based on the average accuracy may be biased by the actual choice of the classes and the available distinctive features. 

Compared to the audio-based reactive classification approach presented in \cite{valada2017}, our approach is outperformed by ~\SIrange{5}{6}{\percent}  (CNN vs. CNN + LSTM), regarding average classification accuracy, but provides a look-ahead in front of the vehicle.

For future work, we will extend the approach to semantically segmented images.
This promises fine grained information about the location of surface patches and thus additional information for the parameterization of control algorithms.
In order to stabilize classification performance on sequences, we will evaluate the addition of LSTM units to the ResNet50 model.

For an application of the proposed CNN model to road friction estimation, the occurring misclassification of \emph{wet asphalt} and \emph{dirt} as \emph{asphalt} is a critical issue, as this can possibly lead to an over-estimated road friction coefficient, which can in turn reduce control performance in critical situations.
For this reason, we will further investigate the features learned by the ResNet50 model in order to resolve misclassification of those classes.

%% file: 08_acknowledgement.tex
We would like to thank our colleague Felix Gr\"un for valuable discussions.